\documentclass[journal]{IEEEtran}
\usepackage{cite}
\usepackage{amsmath,amssymb,amsfonts}
\usepackage{graphicx}
\usepackage{textcomp}

\usepackage{algorithmic}
\usepackage{algorithm}
\usepackage{array}
\usepackage[caption=false,font=normalsize,labelfont=sf,textfont=sf]{subfig}
\usepackage{mathrsfs}
\usepackage{listings}
\usepackage{textcomp}
\usepackage{stfloats}
\usepackage{graphicx}
\usepackage{picinpar}
\usepackage{url}
\usepackage{colortbl}
\usepackage{soul}
\usepackage{multirow}
\usepackage{pifont}
\usepackage{color}
\usepackage{alltt}
\usepackage[hidelinks]{hyperref}
\usepackage{enumerate}
\usepackage{siunitx}
\usepackage{breakurl}
\usepackage{epstopdf}
\usepackage{pbox}
\usepackage{makecell}
\usepackage{diagbox}
\usepackage{threeparttable}
\usepackage{tabularx}
\usepackage{hhline}

\def\BibTeX{{\rm B\kern-.05em{\sc i\kern-.025em b}\kern-.08em
    T\kern-.1667em\lower.7ex\hbox{E}\kern-.125emX}}

\begin{document}

\title{A Novel ViDAR Device With Visual Inertial Encoder Odometry and Reinforcement Learning-Based Active SLAM Method}
%{Zhanhua Xin\orcidlink{0009-0009-7095-0371}, Zhihao Wang\orcidlink{0009-0008-2645-623X}, Shenghao Zhang, Wanchao Chi\orcidlink{0000-0002-0737-1416}, Yan Meng\orcidlink{0000-0001-6248-5427}, Shihan Kong\orcidlink{0000-0002-6714-1313}, Yan Xiong, Chong Zhang, Yuzhen Liu, and Junzhi Yu\orcidlink{0000-0002-6347-572X}
\author{Zhanhua Xin, Zhihao Wang, Shenghao Zhang, Wanchao Chi, Yan Meng, Shihan Kong, Yan Xiong, Chong Zhang, Yuzhen Liu, and Junzhi Yu,~\IEEEmembership{Fellow,~IEEE}
\thanks{This work was supported in part by the Beijing Natural Science Foundation under Grant 2022MQ05, in part by the CIE-Tencent Robotics X Rhino-Bird Focused Research Program under Grant 2022-07, and in part by the National Natural Science Foundation of China under Grant 62203015, Grant 62303020, Grant 62303021, and Grant 62273351. \emph{ (Corresponding author: Junzhi Yu.).}}% <-this % stops a space
\thanks{Zhanhua Xin, Zhihao Wang, Shihan Kong, Yan Xiong, and Junzhi Yu are with the State Key Laboratory for Turbulence and Complex Systems, Department of Advanced Manufacturing and Robotics, College of Engineering, Peking University, Beijing 100871, China (e-mail: xinzhanhua@stu.pku.edu.cn; zh1haowang@stu.pku.edu.cn; kongshihan@pku.edu.cn; x.yan@stu.pku.edu.cn; junzhi.yu@ia.ac.cn).}
\thanks{Shenghao Zhang, Wanchao Chi, Chong Zhang, and Yuzhen Liu are with the Tencent RoboticsX, Shenzhen 518057, China (e-mail: popshzhang@tencent.com; wanchaochi@tencent.com; chongzzhang@tencent.com; rickyyzliu@tencent.com).}
\thanks{Yan Meng is with the College of Information and Electrical Engineering, China Agricultural University, Beijing 100083, China (e-mail: yan.meng@cau.edu.cn).}
}

\maketitle

\begin{abstract}
In the field of multi-sensor fusion for simultaneous localization and mapping (SLAM), monocular cameras and IMUs are widely used to build simple and effective visual-inertial systems. However, limited research has explored the integration of motor-encoder devices to enhance SLAM performance. By incorporating such devices, it is possible to significantly improve active capability and field of view (FOV) with minimal additional cost and structural complexity. This paper proposes a novel visual-inertial-encoder tightly coupled odometry (VIEO) based on a ViDAR (Video Detection and Ranging) device. A ViDAR calibration method is introduced to ensure accurate initialization for VIEO. In addition, a platform motion decoupled active SLAM method based on deep reinforcement learning (DRL) is proposed. Experimental data demonstrate that the proposed ViDAR and the VIEO algorithm significantly increase cross-frame co-visibility relationships compared to its corresponding visual-inertial odometry (VIO) algorithm, improving state estimation accuracy. Additionally, the DRL-based active SLAM algorithm, with the ability to decouple from platform motion, can increase the diversity weight of the feature points and further enhance the VIEO algorithm's performance. The proposed methodology sheds fresh insights into both the updated platform design and decoupled approach of active SLAM systems in complex environments.
\end{abstract}


\section{Introduction}\label{sec:introduction}
\IEEEPARstart{I}{n} recent years, visual odometry (VO) and visual-inertial odometry (VIO) have made significant advancements. They have seen improvements in theoretical understanding and have been widely adopted in various fields, such as autonomous driving and robot navigation \cite{bib:VInavigation1}. To address the scale uncertainty problem, the visual-inertial system, which combines a camera and an inertial measurement unit (IMU), has been extensively studied and experimentally demonstrated \cite{bib:VOreview}. It is known to be a simple, affordable, and efficient solution for odometry sensing \cite{bib:PPFSLAM, bib:VINreview}. Various VO and VIO algorithms have been proposed in the last decade \cite{bib:VO, bib:ORBSLAM2, bib:MSCKF, bib:OKVIS, bib:ROVIO, bib:ORBSLAM3, bib:VINS-Mono, bib:VIOMAV, bib:DDIO}.

Based on VO and VIO, researchers have made continuous efforts to integrate various sensors, to adapt to different simultaneous localization and mapping (SLAM) algorithm application scenarios, or enhance accuracy and robustness. About the legged robots, researchers have developed visual-inertial-leg odometry (VILO) algorithms that utilize a combination of sensors including camera, IMU, leg joint encoders, and contact sensors \cite{bib:legged, bib:VILO}. In the fields of autonomous driving and wheeled robots, researchers have explored the incorporation of encoder data from wheels into a tightly coupled optimization process \cite{bib:VIEO-wheel}. Additionally, to mitigate operational bias during outdoor long-distance navigation, researchers have explored the integration of GPS information into the SLAM framework (VIGO) \cite{bib:VIGO}. For climbing robots on large workpieces, researchers extracted the adsorption constraints on robot motion using CAD model to reduce the drift of VIO \cite{bib:VITO}. In recent years, event cameras have also been incorporated into visual-inertial odometry frameworks, providing high temporal resolution and robustness to challenging lighting conditions\cite{bib:event-VIO}.

Currently, monocular cameras are usually mounted directly on the platform in odometry systems. This constraint results in a restricted field of view (FOV) due to a limited angle range, mechanical obstructions, and constrained installation positions. To enhance the FOV for monocular odometry algorithms, researchers have explored the use of visual or visual-inertial systems mounted on pan-tilt platforms \cite{bib:pan-tilt_scan}. These systems aim to achieve specific tasks or improve odometry algorithm performance. Typically, Yuwen {\it{et al.}} have proposed a Gaze control method \cite{bib:GazeSLAM}, wherein the positioning camera on an Automated Guided Vehicle (AGV) actively controls its gaze to enhance the accuracy of the SLAM algorithm.

However, installing the visual or visual-inertial system on a pan-tilt can introduce additional challenges to the SLAM algorithm. Concretely, some pan-tilt platforms lack slip rings, preventing them from achieving continuous periodic rotation. Moreover, converting the pose of the visual-inertial system equipped on the pan-tilt into the carrying platform becomes problematic. In addition, severe pan-tilt motions will cause intense distortion in the rolling shutter camera. The IMU may also reduce accuracy due to insufficient dynamic response \cite{bib:IMU}, thereby impacting the precision of VIO. In addition, there is a physiological basis in nature, that the visual perception organs and acceleration perception organs are not relatively fixed. Studies have shown that the vestibular organ in the inner ear serves as the posture and acceleration perception organ in human body\cite{bib:vestibular}. However, the eyeball, as a visual perception organ, is capable of free movement.

Inspired by the continuous rotation of mechanical LiDAR and the aforementioned physiological mechanisms, this paper introduces a novel ViDAR (Video Detection and Ranging) system, which integrates a fixed IMU, a motor with an encoder, and a rotating camera. Besides, a calibration method of ViDAR is developed. Building upon this, a tightly coupled VIEO algorithm for the ViDAR is designed. A deep reinforcement learning (DRL) based active SLAM method is proposed to drive the spinning camera for assistance of visual SLAM. ViDAR enables active SLAM algorithms to operate independently of the platform's motion by providing a broader field of view (FOV) and intrinsic rotational capabilities. This decoupling eliminates the requirement for platform-driven motion adjustments, allowing the SLAM algorithm to actively explore and map the environment without imposing specific kinematic constraints on the hosting platform. It should be noted that the encoder here is different from the wheel encoder mentioned before. The VIEO was built upon the VINS-Mono framework \cite{bib:VINS-Mono}.

We also collected visual-inertial-encoder datasets in rosbag format, which are valuable for algorithm testing and further program development. Both the code and datasets have both been made open source on GitHub\footnote{\href{https://github.com/XinZhanhua/VIEO.git}{https://github.com/XinZhanhua/VIEO.git}}.
Accounting for the above issues, the main contributions made in this paper are summarized as follows:
\begin{enumerate}
\item{A ViDAR device that is distinct from the pan-tilt mechanism is proposed and implemented. A ViDAR calibration method based on least squares optimization is proposed, which can calibrate the pose and rotation direction between the IMU, encoder, and camera.}
\item{A tightly coupled odometry algorithm VIEO is proposed, which integrates visual, inertial, and encoder data. A rotational motion freedom is incorporated into the odometry.}
\item{A novel platform-motion-decoupled DRL-based control strategy of the camera's independent rotation is proposed to enhance visual SLAM positioning accuracy.}
\end{enumerate}

The rest of this paper is organized as follows. Section~\ref{sec:Design} is about the mechatronic design and calibration of ViDAR. Section~\ref{sec:Method} introduces the state estimation method based on the VIEO algorithm and the active SLAM method based on reinforcement learning of ViDAR. Section~\ref{sec:Experiments} describes details of the algorithm implementation and presents hardware experiment results. Finally, Section~\ref{sec:Conclusion} gives a brief summary and future work.

\section{Design and Calibration of the ViDAR}\label{sec:Design}
\subsection{Mechatronic Design}
\begin{figure}
\centering
\includegraphics[width=0.44\textwidth]{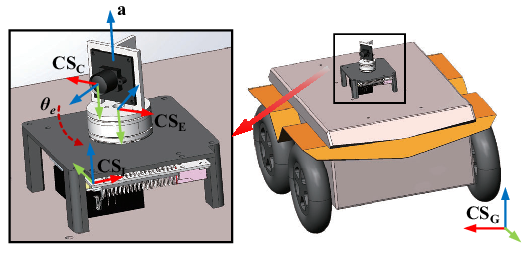}
\caption{Clearpath's jackal car equipped with a ViDAR. The camera coordinate system undergoes a translation and a rotation of angle $\theta_e$ along the rotation axis $\mathbf{a}$ w. r. t. the encoder coordinate system.}
\label{fig:mechanical}
\end{figure}

First, we define the concept of a Visual-Inertial Encoder System (VIENS). Broadly speaking, a VIENS consists of a rotatable CMOS sensor mounted on the motor rotor, an encoder for measuring the motor's rotational motion, and an IMU rigidly attached to the motor stator. More specifically, we have developed a specific design for VIENS. Besides the abovementioned elements, the Nvidia Jetson Xavier NX is installed at the base position to serve as a platform for data collection and storage. An additional RS-Helios-16P LiDAR was installed on the experimental platform to obtain ground truth. For the subsequent experiments, the ViDAR was mounted on a Clearpath's jackal car. The overall structural and design diagram are illustrated in Fig.~\ref{fig:mechanical}.

\begin{table}[htbp]
\centering
\caption{Detailed Specifications of the ViDAR Platform}
\begin{tabular}{ll}
\hline
Items & Characteristics \\ \hline
Dimensions ($L\times W\times H$) & 115 mm$\times$125 mm$\times$82 mm \\
Actuators & BLDC motor (PM3505) \\
Camera & KC-200WQJ \\
IMU & JY931 \\
Encoder & AS5600 \\
LiDAR & RS-Helios-16P \\
Slip ring & Moflon MMC119 \\
Micro-control unit & STM32F103C8T6 \\
Main control platform & Jetson Xavier NX \\
Operating system & Ubuntu 20.04 \& ROS noetic \\
\hline
\end{tabular}
\label{tab:details}
\end{table}

\begin{figure*}
\centering
\includegraphics[width=0.85\textwidth]{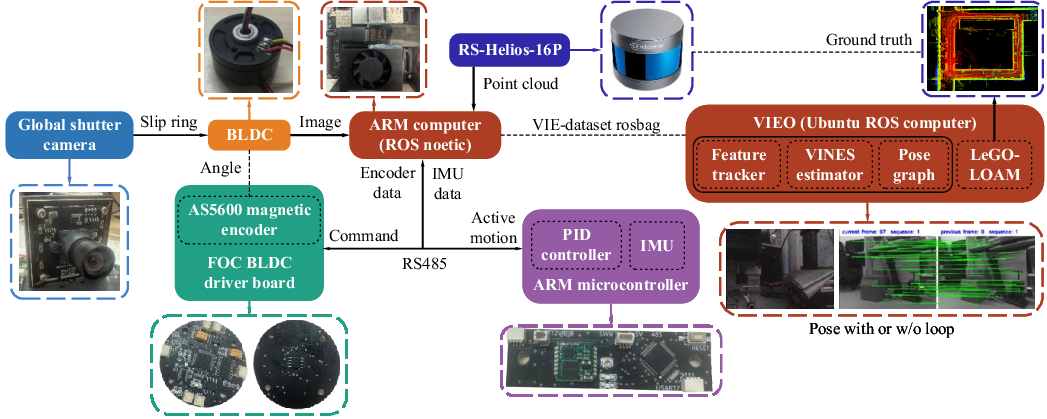} %width=0.9
\caption{An overview of the ViDAR architecture and the proposed VIEO algorithm.}
\label{fig:architecture}
\end{figure*}

\begin{figure*}
\centering
\includegraphics[width=0.86\textwidth]{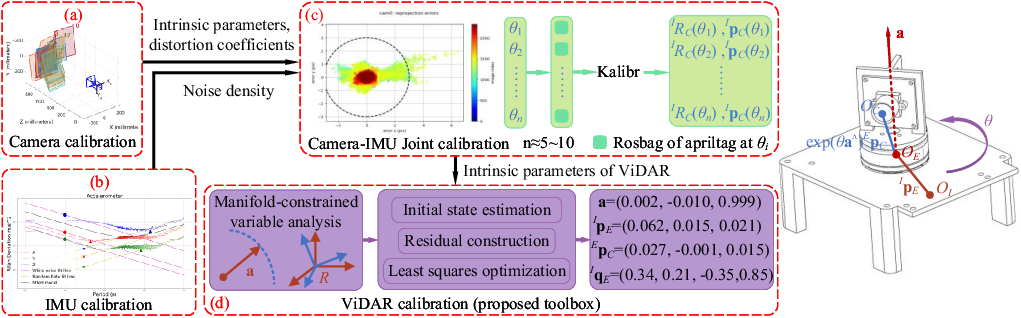}
\caption{Four steps of calibration: (a) camera calibration, (b) IMU calibration, (c) camera-IMU joint calibration, and (d) ViDAR calibration.}
\label{calibration}
\end{figure*}

The driver board and control framework based on SimpleFOC are used to drive the brushless direct current (BLDC) motor in ViDAR. The AS5600 magnetic encoder is utilized as the measurement sensor for the rotor position, which participates in closed-loop control as the actual data input for the position PID controller. Besides, the AS5600 encoder data is outputted via the RS485 bus and collected by the Jetson Xavier NX for tightly coupled optimization in VIEO. More design details are listed in Table~\ref{tab:details}. The overall ViDAR software and hardware architecture diagram is shown in Fig.~\ref{fig:architecture}.

\subsection{ViDAR Calibration}\label{sec:Calibration}
In this paper, lowercase letters, boldface lowercase letters, and uppercase letters are used to denote scalars, vectors, and matrices respectively. When representing rotation matrices and displacement vectors, the superscript on the left denotes the reference coordinate system, while the subscript on the right denotes the target coordinate system. For example, $^IR_C$ denotes the rotation matrix from the IMU coordinate system to the camera coordinate system. The superscript symbol $ \land $ is employed to indicate the mapping of a 3d vector $\mathbf{v}\in \mathbb{R} ^3$ to a three-dimensional skew-symmetric matrix, as follows:
\begin{equation}
\label{eqn:skew-symmetric}
\mathbf{v}^{\land}=\left[ \begin{matrix}
	0&		-v_3&		v_2\\
	v_3&		0&		-v_1\\
	-v_2&		v_1&		0\\
\end{matrix} \right] .
\end{equation}
The special orthogonal group SO(3) can typically be represented by an expansion involving the angle and axis of rotation: $\exp \left( \theta \mathbf{a}^{\land} \right) =\cos \theta I+\left( 1-\cos \theta \right) \mathbf{a}\mathbf{a}^T+\sin \theta \mathbf{a}^{\land}$, where the rotation axis $\mathbf{a}$ belongs to a unit spherical manifold.

\begin{figure}[!t]
\centering
\includegraphics[scale=0.8]{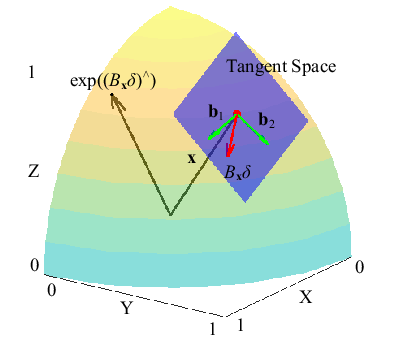}
\caption{The definition of a differential element in a spherical manifold and its addition representation.}
\label{fig:rotate}
\end{figure}

\begin{figure*}
\centering
\includegraphics[scale=0.9]{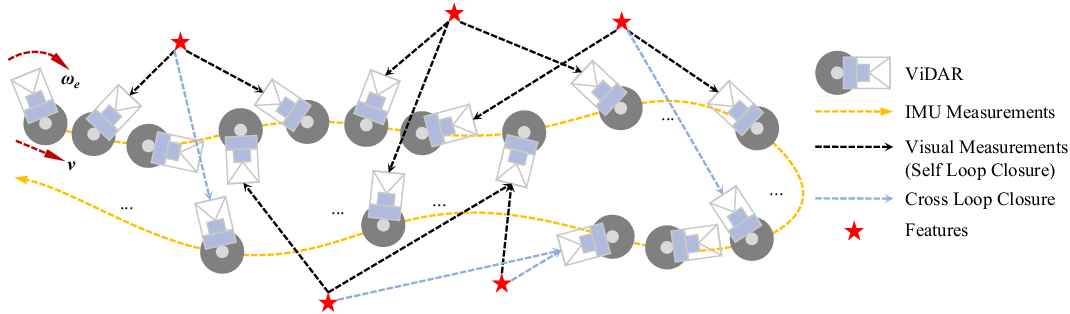}
\caption{ViDAR device and VIEO operation effects schematic. While the ViDAR device runs along a certain trajectory, its camera continuously rotates in cycles under the drive of a motor. The co-visibility relationship depicted in it can be divided into self and cross loop closure.}
\label{fig:visual_radar}
\end{figure*}

Before employing ViDAR for the VIEO, it is imperative to establish the camera's relative pose with the IMU at various angular positions during its rotation. There are four steps in the ViDAR calibration process, as illustrated in Fig.~\ref{calibration}. We employed the Zhang Zhengyou calibration method~\cite{bib:calib_camera}, allan\_variance\_ros toolbox, and Kalibr toolbox~\cite{bib:kalibr} in camera calibration, IMU calibration, and joint calibration respectively. In camera-IMU joint calibration, multiple datasets were collected as the ViDAR fixed to different positions. In the proposed ViDAR calibration toolbox, we used bundle adjustment by the least squares optimization library Ceres to calibrate the state in the ViDAR calibration problem. The residual blocks in this process consist of two types: attitude residuals and position residuals. The state can be defined as follows:

\begin{equation}
\label{eqn:x_calib}
\mathcal{X} _c=\left[ \theta _1, \theta _2, ... \theta _n, \,\mathbf{a}^T, \,^I\mathbf{s}_{C0}^{T}, \,^I\mathbf{p}_{E}^{T}, \,^E\mathbf{p}_{C}^{T} \right] ^T,
\end{equation}
where $\mathbf{a}$ represents the axis of rotation, the rotation matrix $^IR_E$ using the Cayley--Gibbs--Rodriguez (CGR) parameterization\cite{bib:CGR} is expressed as $^I\mathbf{s}_{C0}$, which is a minimal representation. $\theta_i$ represents the angles of the ViDAR for different groups of data in the joint calibration process. $^I\mathbf{p}_E$ represents the position of the encoder coordinate system w. r. t. the IMU coordinate system, while $^E\mathbf{p}_C$ represents the position of the camera coordinate system w. r. t. the encoder coordinate system. Then we get two types of residuals:
\begin{equation}
\label{eqn:calib_residual1}
\mathbf{r}_{1i}\left( \mathbf{z}_i,\mathcal{X} _c \right) = \left( \exp \left( \theta_i \mathbf{a}^{\land} \right) \,^IR_{C0}-\,^IR_C\left( \theta_i \right) \right) \mathbf{p} ,
\end{equation}

\begin{equation}
\label{eqn:calib_residual2}
\mathbf{r}_{2i}\left( \mathbf{z}_i,\mathcal{X} _c \right) = \,^I\mathbf{p}_E+\exp \left( \theta_i \mathbf{a}^{\land} \right) \,^E\mathbf{p}_C-\, ^I\mathbf{p}_C\left( \theta_i \right) ,
\end{equation}
where $\mathbf{p}$ is the projection vector, $z_i$ is the observed value in the joint calibration step of the camera and IMU, including $^IR_C\left( \theta_i \right)$ and $^I\mathbf{p}_C\left( \theta_i \right)$. The optimization problem can be defined as minimizing the sum of the L2 norm of all residuals
\begin{equation}
\label{eqn:calib}
\underset{\mathcal{X} _c}{\min}\left\{ \sum_{i=1}^n{\left( \left\| \mathbf{r}_{1i}\left( \mathbf{z}_i,\mathcal{X} _c \right) \right\| _{P_1}^{2}+\left\| \mathbf{r}_{2i}\left( \mathbf{z}_i,\mathcal{X} _c \right) \right\| _{P_2}^{2} \right)} \right\} .
\end{equation}

The Jacobian matrix of the residuals w. r. t. the variables being optimized can be calculated. A right perturbation model is used to calculate the partial derivative of $^I\mathbf{s}_{E}$, which lies on an SO(3) manifold. The axis $\mathbf{a}$ lies on a unit sphere manifold. The definition of the infinitesimal elements and addition operation on the unit sphere manifold refers to the methods employed in the Manifold ToolKit (MTK) \cite{bib:MTK}. For a unit vector $\left\| \mathbf{x} \right\| _2=\left\| x^2+y^2+z^2 \right\| _2=1$ in the space, the perturbation model is defined as follows:
\begin{equation}
\label{eqn:plus}
\mathbf{x} \boxplus\delta =\,\exp \left( \left( B_\mathbf{x}\delta \right) ^{\land} \right)\mathbf{x},
\end{equation}
where $ \delta $ is a two-dimensional infinitesimal element, $B_\mathbf{x}$ can be expressed as
\begin{align}
\label{eqn:Bx}
\renewcommand\arraystretch{1.2}
B_\mathbf{x}=\left[
\begin{matrix}
	1-{x^2}/{(1+z)}&		-{xy}/{(1+z)}\\
	-{xy}/{(1+z)}&		1-{y^2}/{(1+z)}\\
	-x&		-y\\
\end{matrix} \right] .
\end{align}

Apparently, $\mathbf{x}$ and the two column vectors $\mathbf{b}_1, \mathbf{b}_2$ are orthogonal to each other. Besides, it is easily confirmed that $\left\| B_\mathbf{x} \delta \right\| _2=\delta$. Therefore, if $\delta=\left[\delta_1, \delta_2\right]^T$, consider $B_\mathbf{x} \delta$ as the Lie algebra associated with the special orthogonal group, then its corresponding rotation can be described as a small rotation of angle $\delta$ along the axis $\delta_1\mathbf{b}_1+\delta_2\mathbf{b}_2$, which is orthogonal to $\mathbf{x}$. Applying this rotation to $\mathbf{x}$, we obtain a vector $\mathbf{x} \boxplus\delta$ that still lies on the spherical manifold. A more explicit graphical representation is shown in Fig.~\ref{fig:rotate}.

In addition, the derivative calculation of $\delta$ requires the approximate Baker-Campbell-Hausdorff (BCH) formula, which is used to describe the relationship between small changes in a Lie group and changes in the Lie algebra: $\exp \left( \left( \delta \phi +\phi \right) ^{\land} \right) \approx \exp \left( \phi ^{\land} \right) \exp \left( J_r\left( \phi \right) \delta \phi^{\land} \right)$, where $J_r\left( \theta \mathbf{a} \right)$ can be described as $\left(\sin \theta/\theta\right)I+\left( 1-\sin \theta/\theta \right) \mathbf{aa}^{\mathrm{T}}-\left(\left(1-\cos \theta\right)/\theta\right)\mathbf{a}^{\land}$. By comprehensively utilizing the Rodriguez formula, BCH formula, and first-order Taylor expansion of $\exp$ mapping, the Jacobian matrix can be solved. Therefore, the partial derivative of the residual w. r. t. the variable to be optimized can be expressed.

\section{ViDAR-Based Odometry and Active SLAM}\label{sec:Method}
\subsection{Tightly Coupled Monocular VIEO}\label{sec:VIEO}
\subsubsection{Initialization, Frontend, and Synchronization}

Before the VIEO starts running, an initialization is required. First, the FOC motor driver inside the ViDAR performs FOC initialization to estimate the initial state. Afterward, the VIEO algorithm performs a visual-inertial joint initialization while the ViDAR remains fixed in a specific orientation to estimate scale, gravity acceleration $\mathbf{g}$, velocity, and IMU biases. It aligns the estimated pose information from visual structure from motion (SFM) and the result of IMU preintegration.

The front-end algorithm comprises feature detection, optical flow tracking, and initial state estimation. An initial estimate of feature points is generated based on sensors, which enhances optical flow tracking performance under high FOV motion conditions. The ViDAR completes initialization in stationary mode and enters the normal operating state. Then, a motion pattern can be specified for the ViDAR, as shown in Fig.~\ref{fig:visual_radar}.

\begin{figure}[!t]
\centering
\includegraphics[scale=1]{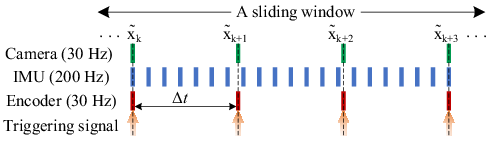}
\caption{Diagram of multi-sensor data in visual-inertial encoder system.}
\label{fig:synchronization}
\end{figure}

In ViDAR, we use the hardware synchronization method for encoder data. A 30~Hz signal trigger is used to interface with the camera and the FOC driver, enabling the acquisition of returned images and encoder data, as shown in Fig.~\ref{fig:synchronization}.

\subsubsection{Optimization Problem}
As for sensor fusion in VIENS, the motor adds extra degrees of freedom, but also introduces more uncertainty. We denote $\mathscr{Z}$ as all measurements, $\mathscr{X}$ as all states, subsets $\mathscr{Z} _{sub}\in \mathscr{Z}$ and $\mathscr{X} _{sub}\in \mathscr{X}$. Then, VIEO constructs a nonlinear least-squares problem to find $\mathscr{X}$ as the solution of
\begin{equation}
\label{eqn:problem}
\min _{\mathscr{X}}\left\{\sum_i\left\|\boldsymbol{r}_i\left(\mathscr{X}_{s u b}, \mathscr{Z}_{s u b}\right)\right\|_{P_i}^2\right\},
\end{equation}
where $\boldsymbol{r}_i$ residuals mentioned in \cite{bib:VINS-Mono}, including prior residuals, IMU preintegration residuals, visual reprojection residuals, and loop-closure residuals. $P_i$ is a weighting matrix that encodes the relative uncertainty in each $\boldsymbol{r}_i$.

The complete state representation must include all state variables in a sliding window and the inverse depth of all correctly matched map points. To simplify the subsequent derivation process, only the state parameters of one frame are retained in the state variables. Therefore, the error-state representation with the addition of encoder angle data and additional extrinsic parameter data is as follows:
\begin{equation}
\renewcommand\arraystretch{1.2}
\label{eqn:x_optimization}
\begin{split}
&\mathbf{\tilde{x}}=\left[ \mathbf{\tilde{x}}_{s}^{T}\mid \,\,\mathbf{\tilde{x}}_{m}^{T} \right]^T= \\
&\left[ ^I\delta \theta _{G}^{T},\mathbf{\tilde{b}}_{g}^{T},\,^G\mathbf{\tilde{v}}_{I}^{T},\mathbf{\tilde{b}}_{a}^{T},\,^G\mathbf{\tilde{p}}_{I}^{T},\,^E\delta \theta _{C}^{T}, \,\,^I\mathbf{\tilde{p}}_{C}^{T},\,\,^G\mathbf{\tilde{p}}_{f_1}^{T}\,\,...^G\mathbf{\tilde{p}}_{f_N}^{T} \right] ^T
\end{split}
\end{equation}
where $\mathbf{\tilde{x}}_{s}$ is the $21 \times 1$ error state corresponding to the sensing platform, and $\mathbf{\tilde{x}}_{m}$ is the $3N \times 1$ error state of the map. $^E\delta \theta _{C}$ represents the attitude error of the camera w. r. t. the encoder, $^I\mathbf{\tilde{p}}_{C}$ represents the position error of the camera w. r. t. the IMU. The linearized continuous-time error-state equation is:
\begin{equation}
\renewcommand\arraystretch{1.2}
\label{eqn:error-state-equation}
\begin{split}
\mathbf{\dot{\tilde{x}}}&=\left[ \begin{matrix}
	\mathbf{F}_s&		0_{21\times 3N}\\
	0_{3N\times 21}&		0_{3N}\\
\end{matrix} \right] \mathbf{\tilde{x}}+\left[ \begin{array}{c}
	\mathbf{G}_s\\
	0_{3N\times 14}\\
\end{array} \right] \mathbf{n}\\
&=\mathbf{F}_c\mathbf{\tilde{x}}+\mathbf{G}_c\mathbf{n},
\end{split}
\end{equation}
where $0_{3N}$ denotes the $3N \times 3N$ matrix of zeros,
$\mathbf{n}=\left[ \begin{matrix}
	\mathbf{n}_{g}^{T}&		\mathbf{n}_{wg}^{T}&		\mathbf{n}_{a}^{T}&		\mathbf{n}_{wa}^{T}&   n_e&    n_{\omega e}\\
\end{matrix} \right] ^T
$ is the system noise of the Angular velocity sensor, accelerometer, and encoder, $\mathbf{F}_s$ is the continuous-time error-state transition matrix corresponding to the sensor platform state, and $\mathbf{G}_s$ is the continuous-time input noise matrix, i.e.,
\begin{equation}
\label{eqn:Fs}
\mathbf{F}_{\mathrm{s}}=\left[ \begin{matrix}
	-\hat{\omega}^{\land}&		-\mathrm{I}_3&		0_3&		0_3&		0_3&  0_3&	0_3\\
	0_3&		0_3&		0_3&		0_3&		0_3&  0_3&	0_3\\
	\mathbf{F}_1&		0_3&		0_3&		\mathbf{F}_2&		0_3&  0_3&	0_3\\
	0_3&		0_3&		0_3&		0_3&		0_3&  0_3&	0_3\\
	0_3&		0_3&		\mathrm{I}_3&		0_3&		0_3&  0_3&	0_3\\
        0_3&		0_3&		0_3&		0_3&		0_3&  0_3&	0_3\\
        0_3&		0_3&		0_3&		0_3&		0_3&  0_3&	-\hat{\omega}_c \mathbf{a}^{\land}
\end{matrix} \right],
\end{equation}

\begin{equation}
\label{eqn:Gs}
\mathbf{G}_{\mathrm{s}}=\left[ \begin{matrix}
	-\mathrm{I}_3&		0_3&		0_3&		0_3&  0_{3\times1}& 0_{3\times1}\\
	0_3&		\mathrm{I}_3&		0_3&		0_3&  0_{3\times1}& 0_{3\times1}\\
	0_3&		0_3&		-C^T\left( \,^I\mathbf{\hat{s}}_G \right)&	0_3&  0_{3\times1}& 0_{3\times1}\\
	0_3&		0_3&		0_3&		\mathrm{I}_3&  0_{3\times1}& 0_{3\times1}\\
	0_3&		0_3&		0_3&		0_3&  0_{3\times1}& 0_{3\times1}\\
        0_3&		0_3&		0_3&		0_3&  \mathbf{g}_1& \mathbf{g}_2\\
        0_3&		0_3&		0_3&		0_3&  0_{3\times1}& \mathbf{a}\\
\end{matrix} \right] ,
\end{equation}
where $\mathbf{F}_1 = -C^T\left( \,^I\mathbf{\hat{s}}_G \right) \,^I\mathbf{\hat{a}}^{\land}$, $\mathbf{F}_2 = -C^T\left( \,^I\mathbf{\hat{s}}_G \right)$, $\mathbf{g}_1 = -\partial/\partial \theta_e(\exp(\theta_e\mathbf{a}^\land)\, J_r\left( \theta_e \mathbf{a} \right))\,\mathbf{a}^\land\,^E\mathbf{p}_C\, \hat{\omega}_e$, $\mathbf{g}_2 = -\partial/\partial \theta_e\exp(\theta_e\mathbf{a}^\land)\,^E\mathbf{p}_C$, $C(\mathbf{s})$ means the rotation matrix of $\mathbf{s}$, $\theta_e$ and $\omega_e$ means the angle and angular velocity of the encoder, the hat means the state estimate propagation model. The fundamental difference between VINS and VIENS in the error-state equation highlights the necessity of developing the VIEO algorithm. State observation refers to the position of the $i$-th landmark in the camera, which is the coordinate description in the camera's normalized coordinate system.
\begin{equation}
\renewcommand\arraystretch{1.1}
\label{eqn:observation1}
\begin{array}{c}
	\left[ \begin{matrix}
	p_x&		p_y&		p_z\\
\end{matrix} \right] ^T=\,^C\mathbf{p}_{f_i}=\exp \left( \theta _e\mathbf{a}^{\land} \right) C\left( \,^{C}\mathbf{s}_{I0} \right)\cdot\\
	 \left( C\left( \,^I\mathbf{s}_G \right) \left( \,^G\mathbf{p}_{f_i}-\,^G\mathbf{p}_I \right) -\,^I\mathbf{p}_C \right),
\end{array}
\end{equation}
where the normalized 2D observation coordinates in the camera's frame $z$ can be described as $\mathbf{z}=1/p_z\left[ \begin{matrix}
	p_x&		p_y\\
\end{matrix} \right] ^T$.

\begin{figure}[!t]
\centering
\includegraphics[scale=0.9]{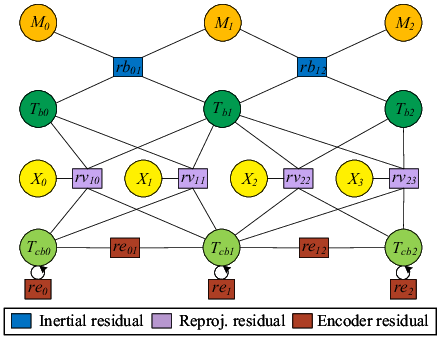}
\caption{ Factor graph representation for VIEO. $M$ represents velocity, angular velocity bias, and acceleration bias. $T_b$ denotes the base pose, $X$ is feature point, and $T_{cb}$ is the pose of the camera w. r. t. IMU.}
\label{fig:factor_graph}
\end{figure}

In particular, to characterize the various types of residuals in equation \eqref{eqn:problem}, the IMU preintegration residuals are the same as in \cite{bib:VINS-Mono}. The visual re-projection is defined as follows:
\begin{equation}
\label{eqn:re-projection2}
\begin{array}{l}
	P_{l}^{cj'}=\,
 ^IR_{Cj}^T\left( ^GR_{Ij}^{T}\left( ^GR_{Ii}\left(^IR_{Ci} \, P_{l}^{ci} \, {\lambda _l^{-1}} \begin{array}{c} \\	\\ \end{array} \right. \right. \right.\\
	\left. \left. \left. \begin{array}{c} \\	\\ \end{array} \, +\,^I\mathbf{p}_{Ci} \right) +\,^G\mathbf{p}_{Ii}-\,^G\mathbf{p}_{Ij} \right) -\,^I\mathbf{p}_{Cj} \right)\\
\end{array}
\end{equation}
where $^IR_{Ci}$ and $^IR_{Cj}$ are the camera pose w. r. t. IMU for two different frames in a sliding window, respectively. $^I\mathbf{p}_{Ci}$ and $^I\mathbf{p}_{Cj}$ are the displacement. The specific derivation process is similar to that in the calibration section. Additionally, encoder residuals can be classified into three types, representing the constraints between the camera's pose w. r. t. the IMU and the encoder angles in ViDAR:
\begin{equation}
\label{eqn:encoder_residual1}
\boldsymbol{r}_{ei1}=\log \left( ^IR_{Ci}\,\,^IR_{E}^{T} \right) -\theta _{ei} \mathbf{a},
\end{equation}

\begin{equation}
\label{eqn:encoder_residual2}
\boldsymbol{r}_{ei2}=\exp \left( \theta _{ei}\mathbf{a}^{\land} \right)\, ^E\mathbf{p}_C+\,^I\mathbf{p}_E-\,^I\mathbf{p}_{Ci},
\end{equation}

\begin{equation}
\label{eqn:encoder_residual3}
\boldsymbol{r}_{eij}=\varphi \left( \, ^IR_{Cj} \, \right)-\varphi \left( \, ^IR_{Ci} \, \right) -\left( \theta _{ej}-\theta _{ei} \right),
\end{equation}
where $ \varphi \left( \right) $ refers to the equation used to calculate the rotation angle based on the rotation matrix. We use the automatic differentiation feature provided by the Ceres library to solve for the Jacobian matrix of the encoder residuals. The optimization problem of the VIEO algorithm can be simplified into a factor graph as shown in Fig.~\ref{fig:factor_graph}.

\subsubsection{Cross-Frame Co-visibility Relationships}
The loop detection method in the VIEO algorithm is implemented using the bag-of-words approach based on the DBoW2 library, which is the same as in VINS-Mono. When the ViDAR enters into reciprocating or rotating mode, a large number of cross-frame co-visibility relationships emerge. On one hand, self-loop closure might be triggered at a nearby angle after the ViDAR completes one motion cycle. On the other hand, if a vehicle equipped with ViDAR moves back and forth along a certain path but in different directions, the ViDAR is also prone to triggering cross-loop closure with frames from various previous fields of view. The effect is shown in Fig.~\ref{fig:visual_radar}.

\subsection{DRL-Based Active SLAM}
\subsubsection{System Overview}
\begin{figure}
\centering
\includegraphics[scale=0.62]{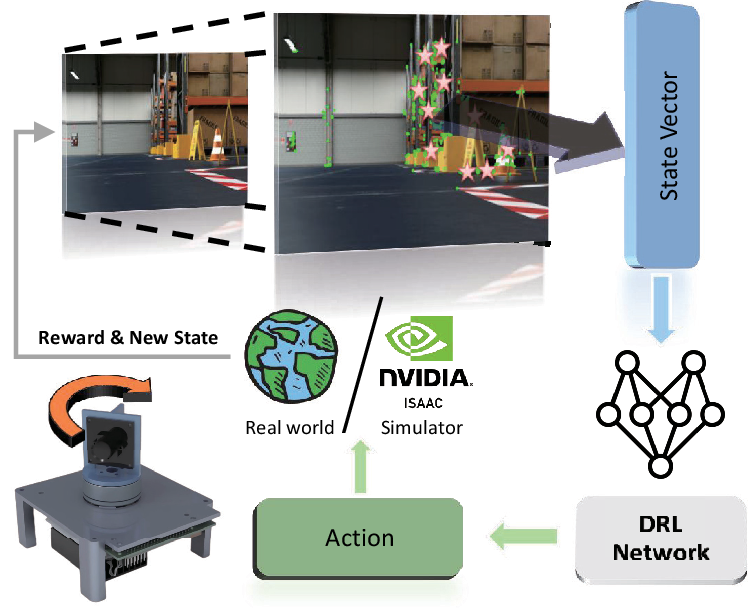}
\caption{The pipeline of DRL-based control approach for ViDAR.}
\label{fig:rl_framwork}
\end{figure}

The motion flexibility of ViDAR allows us to design specific rules enabling it to operate in a defined motion pattern. Thus, we proposed a DRL-based active SLAM method suitable for ViDAR, as shown in Fig.~\ref{fig:rl_framwork}. Note that unlike the traditional active SLAM algorithm, which aims at autonomously moving vehicles around to get environmental information as much as possible. Here active SLAM means ViDAR can control its rotation in autonomy, thus achieving higher system precision, while keeping the vehicle's original trajectory unchanged.

Concretely, images are pre-processed using a novel feature points distribution metric. The resulting observation vector and the current rotation speed are then used as the input. Then rewards are retrieved to guide for better performance. We choose to use the SAC (Soft Actor-Critic) \cite{bib:sac} algorithm due to its high efficiency in exploiting historical data. We have trained the above DRL-based control pattern on the Nvidia Isaac Sim simulator with various scenarios and trajectories.

\subsubsection{DRL Problem Definition}
During the agent's interactions with the simulated environments, the period can be seen as a Markov Decision Process (MDP), which can be represented with the equation $\mathcal{M}=\langle \mathcal{S},\mathcal{A},\mathcal{P},r,\mathcal{\gamma} \rangle$. $\mathcal{S}$ means state space. $\mathcal{A}$ is action space, meaning the camera's spinning speed. $\mathcal{P}(s_{t+1} | s_{t}, a_{t})$ is the transition probability from the current state and action to the new state, which is decided by the environment. $r(s_{t},a_{t})$ is the reward function, which can be calculated from the state and action. $\gamma \in [0,1]$ is the discount factor which helps the agent to focus more on current rewards instead of the distant ones. Thus for a DRL agent, its ultimate goal is to maximize the return $\mathcal{R}$ during interactions with the environment in each episode. The return is formulated as
\begin{equation}
\label{equ_1}
\mathcal{R}=\sum_{t=0}^{n} \gamma ^{t} r(s_{t},a_{t})
\end{equation}

\subsubsection{Metric of Cross-images Feature Points Distribution Uniformity}
Empirically, feature points with higher distribution uniformity are considered superior. Unlike norm SLAM extracting feature points from given images passively, we design a metric to judge this uniformity and flexibly choose the better view. Drawing inspiration from both the Shannon diversity index \cite{bib:shannon} and the ORB-SLAM series, we first divide the raw image into 20 columns and 20 rows, then assign every small grid as a kind of species. The quantity of feature points detected within a grid is viewed as the number of related species with a proportion $p_i$. And $n_{fp}$ represents for total quantity of feature points detected within the whole image. We obtain the metric $U$ for judging cross-image feature point distribution uniformity as

\begin{equation}
\label{equ:uniformity_metric}
U=\frac{-\sum_i^Np_i\ln p_i}{\ln n_{fp}}
\end{equation}

\subsubsection{Action, State, and Reward}
We choose to use a continuous representation for the action space. For every action predicted by the agent $a_t\in[-1.0,1.0]$. The action is performed like a rotation speed which is either clockwise or anticlockwise. To run the DRL-based control approach in real time, we meticulously design a state vector. First, the quantity of feature points within every aforementioned grid $n_i$ is included, together with current action $a_{current}$ and the difference between current and last action $a_{diff}$. Then, we employ a strategy to provide the agent with a global view. We use the K-Means++ method to obtain $\theta$ centroids of all feature points but only use the top $k$ centroids with most points related. The overall state vector is depicted as

\begin{equation}
\label{equ:kmeans}
X_{i} = \sum_i^N n_i \cup a_{current} \cup a_{diff} \cup F_{K}(n_{fp}).top(k)
\end{equation}

The reward function is designed with expert knowledge from both DRL and visual SLAM. We design the reward $r_t$ suitable for active SLAM control, which can be depicted as
\begin{equation}
\label{equ:overall_reward}
r_t = r_{q}\cdot r_{u} + r_{a}\cdot r_{d}
\end{equation}
where $r_q$ is the quantity reward to extract more feature points. $r_u$ is the measurement of uniformity as equation~\ref{equ:uniformity_metric}. $r_a$ aims to prevent from stationary and non-exploratory patterns. $r_d$ is used to maintain consistency in output actions. The detailed composition of rewards is
\begin{equation}
\begin{split}
\label{equ:all_rewards}
&r_{q}= (e^{n_{fp}}-e^{\beta n_{fp}})\cdot (e^{n_{fp}}+e^{\beta n_{fp}})^{-1}\\
&r_{u}= U=-(\sum_i^Np_i\ln p_i\,) / \ln n_{fp} \\
&r_{a} = ({\lambda_{a}\sqrt{2\pi}})^{-1}\cdot \exp[-(a\mu_{a})^2\cdot(2\sigma_{a})^{-1}]\\
&r_{d} = ({\lambda_{d}\sqrt{2\pi}})^{-1}\cdot \exp[-(a\mu_{d})^2\cdot(2\sigma_{d})^{-1}]\\
\end{split}
\end{equation}
where $\beta$, $\lambda$ and $\sigma$ are all fixed parameters defined manually before training, to obtain proper reward signals.

\begin{figure}[!t]
\centering
\includegraphics[scale=0.88]{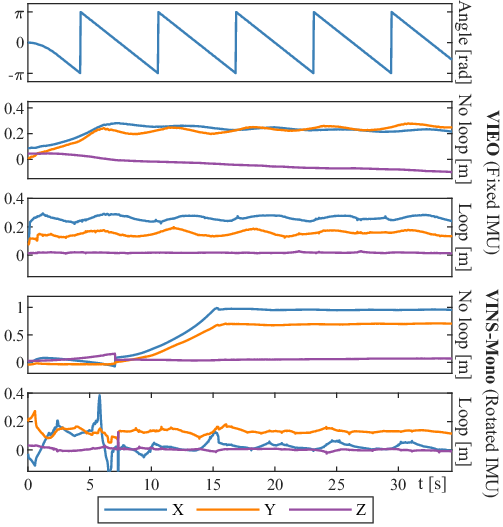}
\caption{The jackal car was fixed in place, while the ViDAR continuously rotated in cycles. The encoder angle and the coordinate position were obtained from VIEO and VINS-Mono.}
\label{fig:experiment24}
\end{figure}

\begin{figure}[!t]
\centering
\includegraphics[scale=0.92]{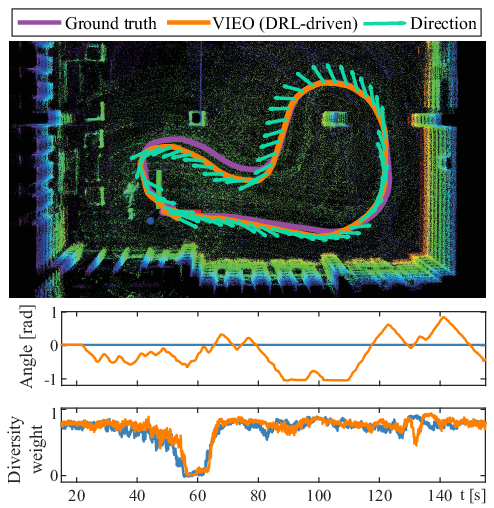}
\caption{ViDAR adjusts the field of view direction driven by agents to improve the diversity weight of feature points.}
\label{fig:experiment130}
\end{figure}

\begin{table}[!t]
    \centering
    \caption{VIEO Algorithm Small Scale Experimental Results}
    \label{tab:result}
    \resizebox{\linewidth}{!}{
    \begin{threeparttable}
        \begin{tabular}{|c|c|c|c|c|}
        \hline
            \textbf{\diagbox{Seq.}{Experiments}} & \textbf{\makecell{VIEO\\(no loop)}} & \textbf{\makecell{VIEO\\(loop)}} & \textbf{\makecell{VINS\\-Mono}} & \textbf{\makecell{Ablation\\studies}} \\ \hline
            \textbf{Stationary (indoor)} & 2.11\% & \textbf{1.65\%} & 2.01\% & 2.57\% \\ \hline
            \textbf{Stationary (outdoor)} & 3.34\% & \textbf{2.11\%} & 2.79\% & X \\ \hline
            \textbf{Reciprocating (indoor)} & 2.16\% & \textbf{1.01\%} & 1.81\%* & 1.53\% \\ \hline
            \textbf{Reciprocating (outdoor)} & 2.50\% & \textbf{0.96\%} & 2.21\%* & 1.45\% \\ \hline
            \textbf{Cyclical (indoor)} & 2.20\% & \textbf{1.05\%} & 1.20\%* & X \\ \hline
            \textbf{Cyclical (outdoor)} & 2.54\% & 0.89\% & \textbf{0.54\%*} & X \\ \hline
            \textbf{DRL-driven (indoor)} & 1.54\% & \textbf{0.87\%} & 1.59\%* & 2.83\% \\ \hline
            \textbf{DRL-driven (outdoor)} & 2.68\% & \textbf{2.32\%} & 2.52\%* & X \\ \hline
        \end{tabular}
        \begin{tablenotes}
            \footnotesize
            \item[1] The datasets are approximately 370 meters long and last over 24 minutes.
            \item[2] * means the dataset was obtained by applying VIENS degradation to VINS while keeping ViDAR fixed and calculating the drift rate in the same route.
            \item[3] X means the dataset cannot be run properly without optimizing ViDAR extrinsic parameters.
        \end{tablenotes}
    \end{threeparttable}}
\end{table}

\section{Experiments}\label{sec:Experiments}
\subsection{Dataset and Experimental Setup}
In the field of SLAM, open-source datasets play a significant role in advancing related new technologies. However, as far as we know, there are currently no existing open-source datasets compatible with the algorithm proposed in this paper. Therefore, we have collected a large number of datasets based on the platform we built and made it publicly available for testing the algorithm proposed in this paper or for further development purposes. The datasets were recorded in rosbag format and included rostopics corresponding to real-time data captured by the camera, IMU, and encoder. Some of the datasets contain scanning data from LiDAR, which can be used to calculate the ground truth of the state.

Within the provided datasets are indoor and outdoor datasets captured by the ViDAR in different motion modes including keeping stationary, uniform reciprocating motion, continuous cyclical rotation, and DRL-based mode. In reciprocating motion mode, the maximum angle is 90 degrees and the swinging period is 10~s. In rotation mode, the speed is set to 10 rpm. In both the fixed and reciprocating motion modes, the motor in the ViDAR uses FOC-based position closed-loop PID control. In the rotational mode, the motor utilizes FOC-based velocity closed-loop PID control. In the stationary mode of ViDAR, the VIENS degenerates to VINS in terms of architecture.

\subsection{Results}
We verified the operational effectiveness of the VIEO algorithm when the car is stationary and the ViDAR is continuously rotating. Two IMUs are fixed on the stator and rotor ends of the motor, respectively. In theory, the estimated position should remain unchanged. The result is shown in Fig.~\ref{fig:experiment24}. With and without loop closure incorporated, the root mean square errors (RMSE) are 0.261~m and 0.026~m in VIEO, and 0.512~m and 0.031~m in VINS-Mono, respectively. It is evident that the self-loop closure of the ViDAR significantly mitigates the position drift in the VIEO algorithm. Furthermore, significant drift was observed in the VINS-Mono system, which operates on the VINS comprising a camera and an IMU mounted on the rotor. This demonstrates that the intense motion of the IMU may degrade the accuracy of state estimation.

\begin{figure*}[!t]
\centering
\includegraphics[width=1\textwidth]{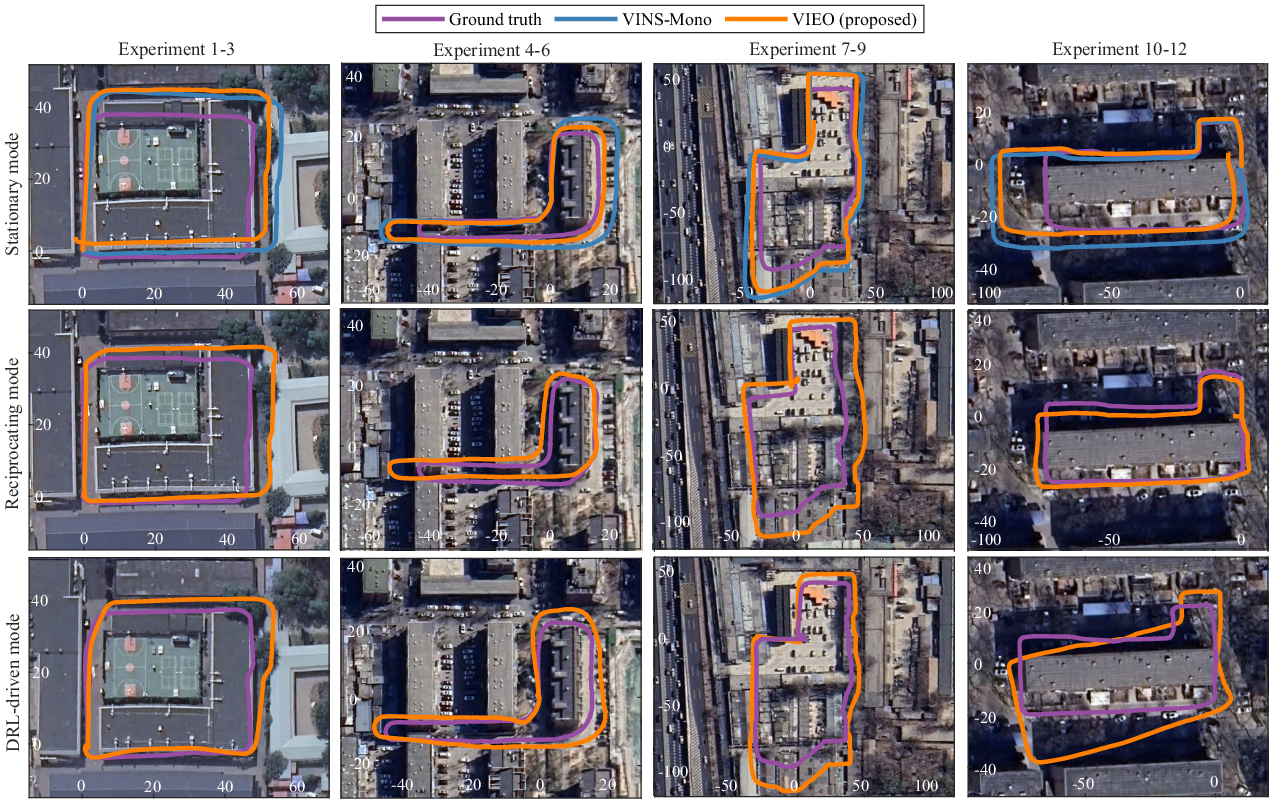}
\caption{Outdoor large-scale real-world experiments with ground truth. Each column represents a set of experiments, in which the Jackal car follows nearly identical trajectories. Under the same conditions, the ViDAR operates in stationary, reciprocating, and DRL-based modes.}
\label{fig:experiment}
\end{figure*}

\begin{figure*}[!t]
\centering
\includegraphics[width=1\textwidth]{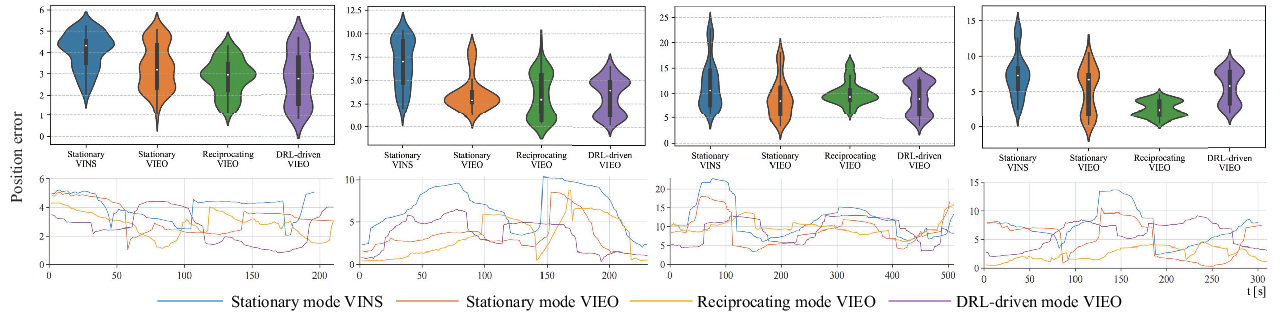}
\caption{Position error of the outdoor large-scale experiments.}
\label{fig:experiment_comparison}
\end{figure*}

\begin{table}[!t]
    \centering
    \caption{VIEO Algorithm Large-Scale Outdoor Experiment RMSE (m)}
    \label{tab:large}
    \begin{threeparttable}
        \begin{tabular}{|c|c|c|c|c|c|}
        \hline
            \textbf{\makecell{ \\Seq.}} & \textbf{\makecell{VINS-\\Mono}} & \textbf{\makecell{ORB-\\SLAM3}} & \textbf{\makecell{DM-\\VIO}} & \textbf{VIEO} & \textbf{\makecell{Length\\(m)}} \\ \hline
            \textbf{Exp 1 (S)} & 4.15 & 3.44 & \textbf{1.81} & 3.48 & 169.10 \\ \hline
            \textbf{Exp 2 (R)} & X & X & X & \textbf{2.97} & 170.04 \\ \hline
            \textbf{Exp 3 (D)} & X & X & X & 3.13 & 162.42 \\ \hhline{|=|=|=|=|=|=|}
            \textbf{Exp 4 (S)} & 7.25 & 7.82 & \textbf{3.94} & 4.03 & 188.58 \\ \hline
            \textbf{Exp 5 (R)} & X & X & X & 4.01 & 188.42 \\\hline
            \textbf{Exp 6 (D)} & X & X & X & \textbf{3.87} & 206.79 \\ \hhline{|=|=|=|=|=|=|}
            \textbf{Exp 7 (S)} & 12.67 & 15.58 & \textbf{7.99} & 9.78 & 389.92 \\ \hline
            \textbf{Exp 8 (R)} & X & X & X & 10.35 & 397.42 \\ \hline
            \textbf{Exp 9 (D)} & X & X & X & \textbf{9.44} & 454.45 \\ \hhline{|=|=|=|=|=|=|}
            \textbf{Exp 10 (S)} & 7.86 & 18.36 & \textbf{3.95} & 5.99 & 225.74 \\ \hline
            \textbf{Exp 11 (R)} & X & X & X & \textbf{2.76} & 229.55\\ \hline
            \textbf{Exp 12 (D)} & X & X & X & 6.06 & 241.43 \\ \hline
        \end{tabular}
        \begin{tablenotes}
            \footnotesize
            \item[1] The RMSE in the table only calculates the absolute trajectory error (ATE) between $XY$ coordinates and the ground truth.
            \item[2] S, R, and D represent the three modes of ViDAR in stationary, reciprocating, and DRL-driven, respectively.
            \item[3] X means the dataset cannot be run properly.
        \end{tablenotes}
    \end{threeparttable}
\end{table}

\begin{figure}
\centering
\includegraphics[scale=0.88]{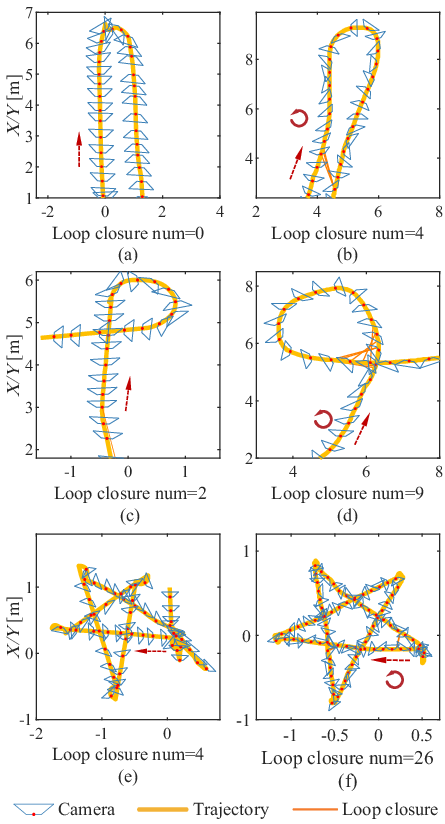}
\caption{ViDAR panoramic view experiment. ViDAR was set in the stationary mode in (a), (c), and (e), and was set in cyclical mode in (b), (d), and (f).}
\label{fig:experiment81}
\end{figure}

Various small-scale indoor and outdoor experiments were conducted under different environmental conditions for multiple ViDAR motion states. The drift was organized into a table as shown in Table~\ref{tab:result}. We provide the drift rate computed by the VINS-Mono algorithm for the corresponding dataset in Table~\ref{tab:result}. The drift rate of the VIEO algorithm varies significantly when the ViDAR operates in different motion modes. In the case of reciprocating mode, the VIEO algorithm benefits from hardware-level co-visibility compensation, resulting in better accuracy. Notably, any datasets collected by ViDAR in non-fixed mode failed to run successfully in VINS-Mono. To address this, we conducted another round of data collection with ViDAR operating in fixed mode along the same trajectory, specifically for comparison experiments. We visualized the results of indoor experiments conducted under the DRL-driven mode, as shown in Fig.~\ref{fig:experiment130}. Compared to the fixed mode, there was a 3.42\% decrease in feature points' diversity weight.

We then conducted a series of large-scale outdoor real-world experiments. The ViDAR was set to stationary mode and uniform reciprocating mode. We utilized a 16-line LiDAR to capture point cloud data simultaneously. Due to the block of view by LiDAR and other parts of the experiment platform, there was no experiment for the cyclical mode. Subsequently, we ran the VIEO algorithm and LeGO-LOAM algorithm offline to generate trajectory outputs and ground truth \cite{bib:LeGO-LOAM}. The results are shown in Fig.~\ref{fig:experiment}. We used EVO tools to evaluate and compare VIEO \cite{bib:evo}. We also present a comparison of the positional errors from the above experiments in Fig.~\ref{fig:experiment_comparison}. The $XY$ drift RMSE is demonstrated in Table~\ref{tab:large}. The results indicate that the VIEO algorithm outperforms VINS-Mono in the stationary mode and further enhances state estimation accuracy when the ViDAR operates in a reciprocating mode or DRL-driven mode. Although more advanced VIO algorithms, such as DM-VIO~\cite{bib:dmvio}, achieve optimal state estimation in ViDAR's stationary mode, the VIEO algorithm, developed based on VINS-Mono, can achieve comparable or even superior results when ViDAR operates in reciprocating or DRL-driven modes.

Besides, we demonstrate the advantages that the additional degrees of freedom in the ViDAR can bring to the VIEO algorithm in certain scenarios. In the scenes depicted in Fig.~\ref{fig:experiment81}, a car equipped with ViDAR underwent various simple, short-range movements, including motion with back-and-forth, spiral, and a pentagram-shaped trajectory. The higher number of loop triggers reflects the panoramic view advantage of ViDAR and VIEO algorithm compared to traditional VIO.

In addition, this involves the necessity analysis of VIEO algorithm ablation experiments and the additional complexity compared to the VIO algorithm. We disabled the optimization of the camera's pose w. r. t. the IMU in the VIEO algorithm, using the trajectory drift rate as the result of the ablation experiment. Optimizing the extrinsic parameters becomes necessary to improve accuracy in VIEO. We conducted a series of ablative experiments, as shown in the fifth column of Table~\ref{tab:result}, without performing extrinsic parameters optimization.

\subsection{Discussion}
The ViDAR, VIEO algorithms, and active SLAM method proposed in this paper demonstrate exciting potential, warranting further in-depth exploration. For example, it is difficult to install multiple sensors for redundancy due to size and computational limitations on lightweight platforms like quadruped robots. In such cases, the VIEO algorithm may have more application potential. Besides, ViDAR has a profound advantage over gimbal. The ViDAR can simultaneously handle active exploration and the state estimation of the carrying platform. In contrast, if a gimbal engages in active exploration, it will cause changes in the camera's pose, thereby altering the state estimation. Moreover, the ViDAR and VIEO algorithms demonstrate strong portability. Integrating a BLDC and encoder can transform any visual-inertial system into a VIENS. Other VIO algorithms built on optimization-based methods can be adapted into a corresponding VIEO algorithm by incorporating encoder residuals and other relevant components.

However, there are still many aspects of this work that remain incomplete or require refinement. The current version of the VIEO algorithm has not fully exploited the advantages of this novel hardware. A considerable portion of co-visible relationships has not been effectively captured. Cross-loop detection under large angular deviations is also a challenging problem. The scarcity of datasets is another challenge faced in this work. Due to the uniqueness of the dataset used for validating the VIEO algorithm, no similar open-source datasets are available online. Regarding the timestamp synchronization issue, we did not incorporate the timestamp discrepancies between the encoder and camera data into the optimization problem due to computational complexity constraints. Instead, we opted for a simpler hardware synchronization approach.

Several aspects of the experimental setup and results presentation merit clarification. Firstly, due to the fairness and the lack of practical significance, this study does not include comparisons of VIEO with fisheye camera-based VIO. VIEO represents an exploratory effort incorporating new sensors. Its value lies more in enhancing the performance of the corresponding VIO algorithms and improving platform motion decoupling through the integration of coupled active SLAM algorithms. Secondly, in the small-scale experiments mentioned above, the VIEO algorithm in cyclical mode did not exhibit superior accuracy compared to VINS-Mono. This issue might be caused by insufficient ViDAR calibration or low-quality feature points within the rotation cycle. Besides, in large-scale experiments, the performance of DRL-based motion mode is not always superior to that of reciprocating motion mode. In certain scenarios, the reciprocating motion of the camera might allow for a broader field of view and a greater number of high-quality feature points within a sliding window, potentially outperforming the DRL-driven mode with more evenly distributed feature points.

\section{Conclusion and Future Work}\label{sec:Conclusion}
The ViDAR design and calibration method proposed in this paper, along with the VIEO algorithm and the DRL-based active SLAM, represents a novel attempt to enhance the performance of SLAM by coupling rotational freedom to the system. The paper presents theoretical principles of the visual-inertial encoder state estimator, releases the open-source implementation of the odometry algorithm, and shares datasets for testing and future development. The algorithm testing on the published datasets shows that the VIEO algorithm can effectively reduce mapping drift in various ViDAR motion patterns. The active SLAM method based on ViDAR can further improve the performance of the VIEO algorithm while decoupling from the platform's motion.

The VINS framework partially limits the ability of the VIEO algorithm. Future developments may consider adopting the ORB-SLAM framework. A more robust feature point detecting and matching strategy and a local map data structure is more suitable for the characteristics of ViDAR. Besides, the optimization-based method for estimating timestamp errors could be explored in future work. In addition, it is necessary to further address the potential visual distortion and motion blur issue that may occur with ViDAR during higher-speed rotation. Possible solutions include running deblurring algorithms or utilizing global shutter cameras with higher frame rates.

% References

%\bibliographystyle{Bibliography/IEEEtranTIE}
%\bibliography{Bibliography/IEEEabrv,Bibliography/BIB_xx-TIE-xxxx}\ %IEEEabrv instead of IEEEfull

\vspace{-1cm}
\begin{IEEEbiography}[{\includegraphics[width=1in,height=1.25in,clip,keepaspectratio]{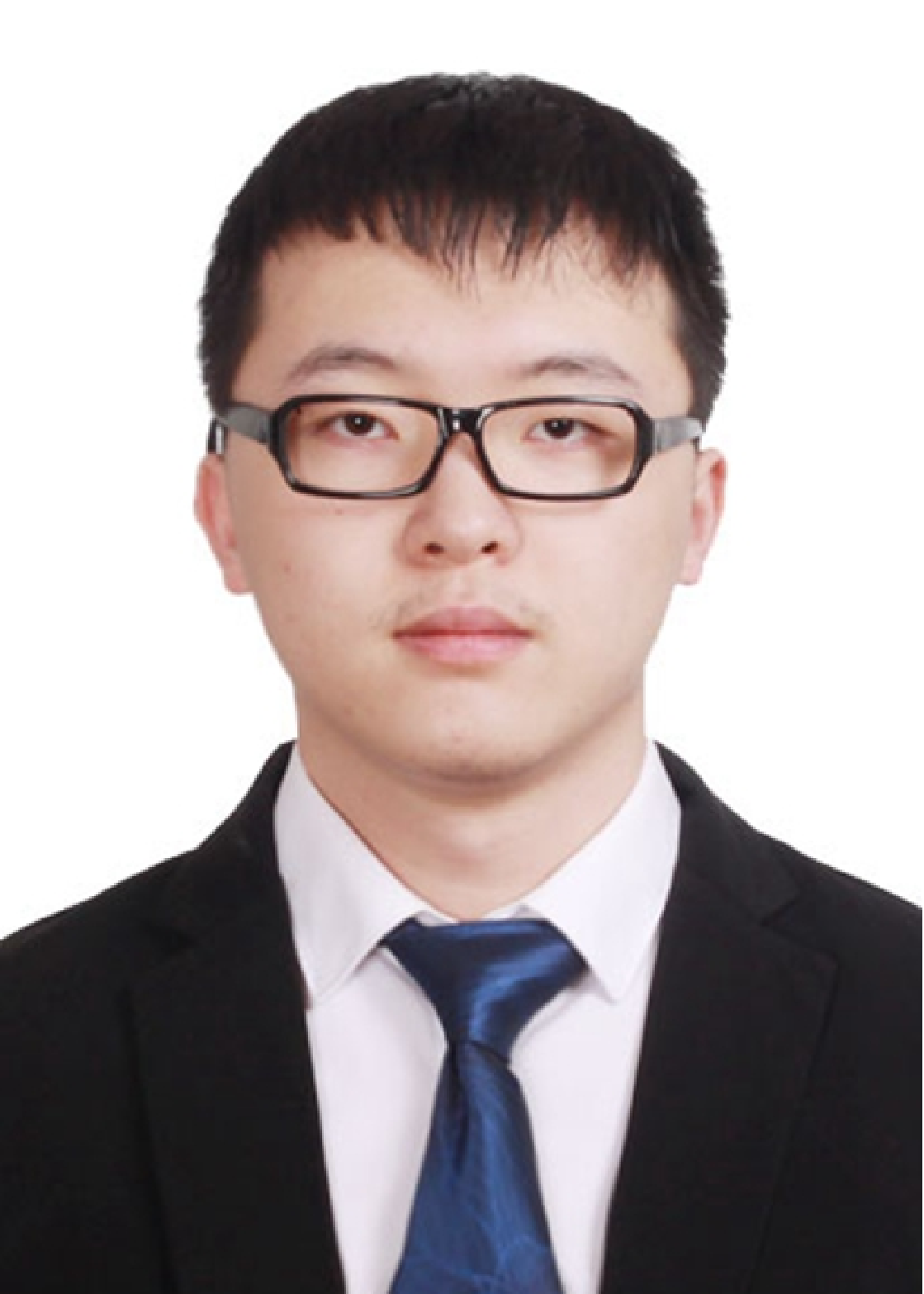}}]
{Zhanhua Xin}
received the B.E. degree in mechatronic engineering from the School of Xuteli, Beijing Institute of Technology, Beijing, China, in 2022. He is currently pursuing the Ph.D. degree in College of Engineering, Peking University, Beijing. His research interests include computer vision, bioinspired underwater robots, and spherical motor.
\end{IEEEbiography}

\vspace{-1cm}
\begin{IEEEbiography}[{\includegraphics[width=1in,height=1.25in,clip,keepaspectratio]{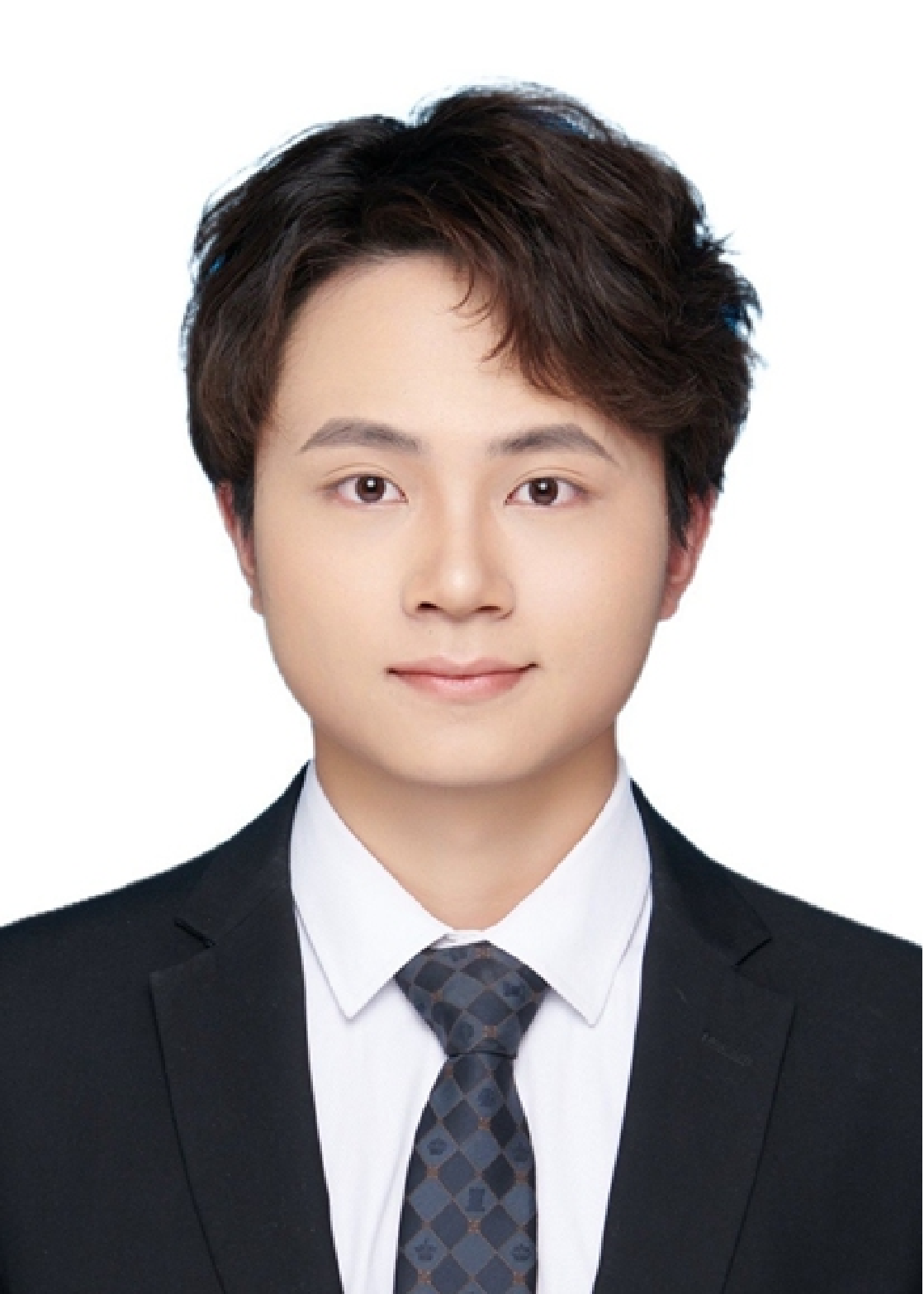}}]
{Zhihao Wang}
received the B.E. degree in Intelligent Automotive Engineering with School of Automotive Engineering, Harbin Institute of Technology, China in 2024. He is currently working toward the M.E. degree in the department of Advanced Manufacturing and Robotics, College of Engineering at Peking University, China. His research interests include embodied intelligence, reinforcement learning and multimodal large language models.
\end{IEEEbiography}

\vspace{-1cm}
\begin{IEEEbiography}[{\includegraphics[width=1in,height=1.25in,clip,keepaspectratio]{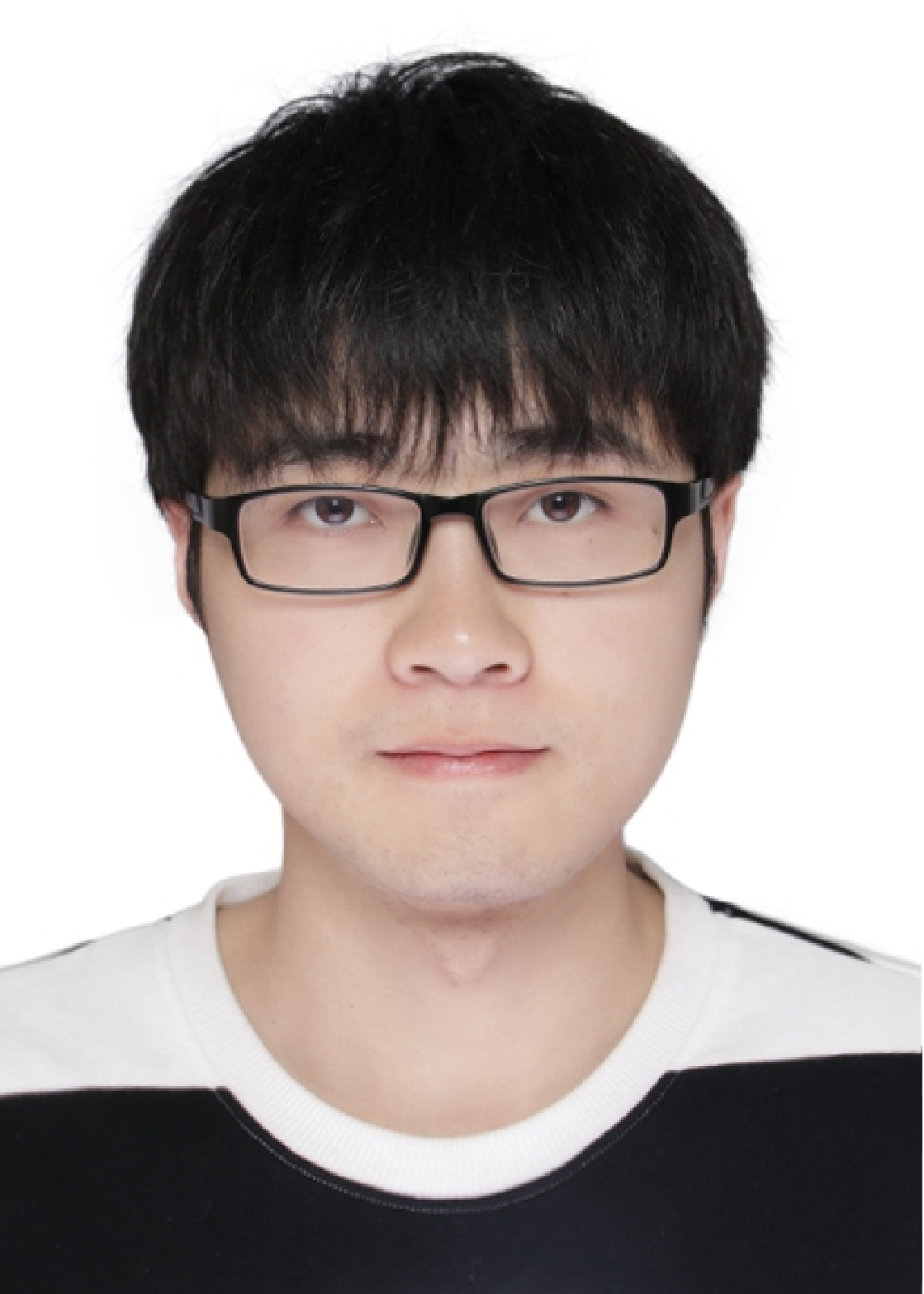}}]
{Shenghao Zhang}
received the M.E. degree in computer application technology from Peking University in 2019. His research scope covered video codec, VR video generation, and SLAM (Simultaneous localization and mapping). He joined Tencent Robotics X in 2019, and his current research interests include SLAM and 3D reconstruction.
\end{IEEEbiography}

\vspace{-1cm}
\begin{IEEEbiography}[{\includegraphics[width=1in,height=1.25in,clip,keepaspectratio]{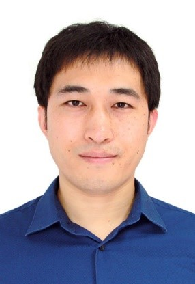}}]
{Wanchao Chi}
received the B.E. degree in Aircraft Design and Engineering and M.E. degree in Engineering Mechanics from Nanjing University of Aeronautics and Astronautics in 2009 and 2011, respectively, and the Ph.D. degree in Mechatronics and Design from Nanyang Technological University (NTU), Singapore in 2016. His research scope covered biomimetics, mechanical design, perception, path planning, collision avoidance, and dynamics modelling and control of unmanned aerial vehicles (UAVs). He joined Tencent Robotics X in 2018, and his current research interests include dynamics modelling and control, motion planning, and state estimation.
\end{IEEEbiography}

\vspace{-1cm}
\begin{IEEEbiography}[{\includegraphics[width=1in,height=1.25in,clip,keepaspectratio]{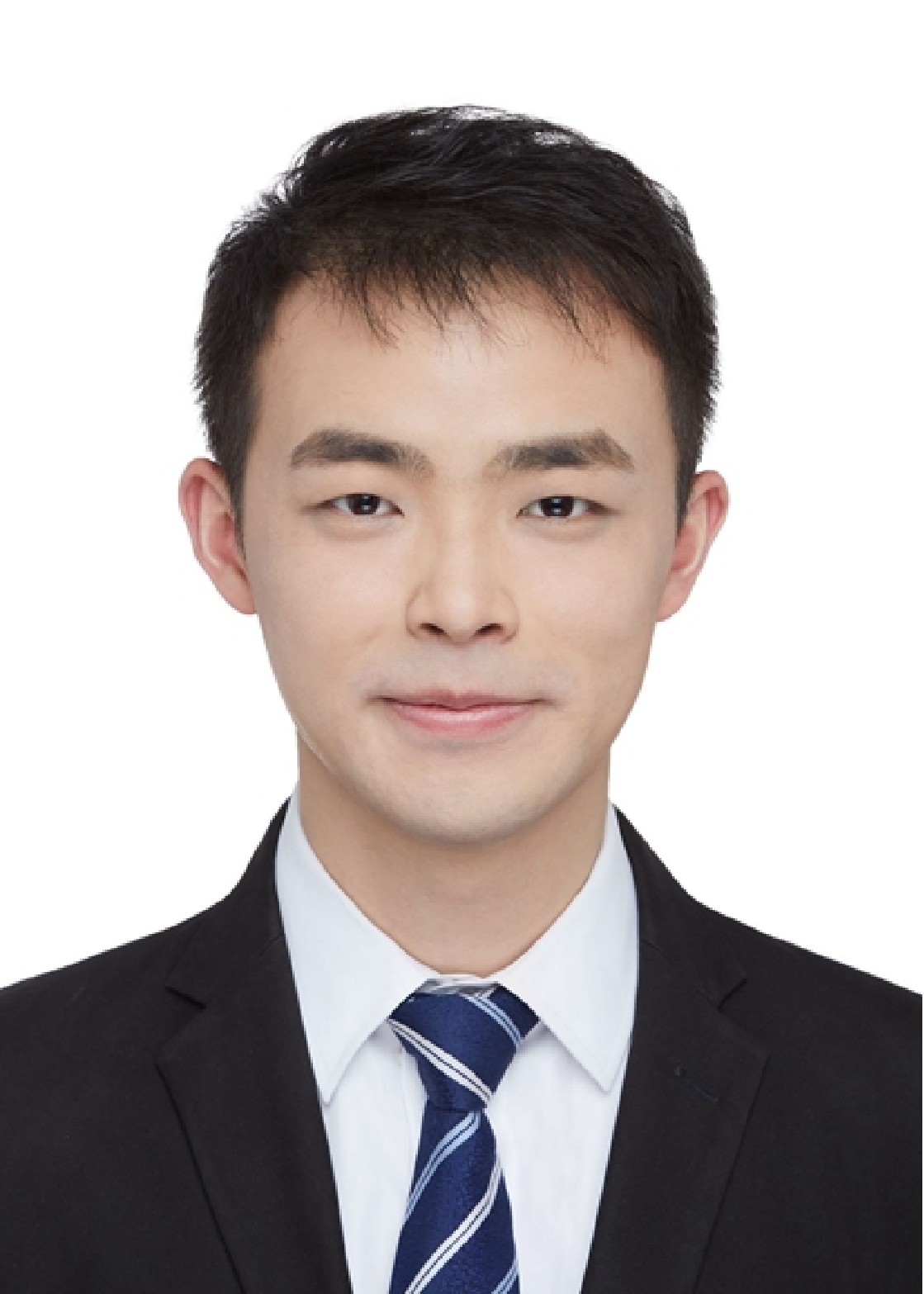}}]{Yan Meng}
received a B.E. degree in mechanical engineering from the School of Mechanical Engineering, University of Science and Technology Beijing, Beijing, China, in 2017 and a Ph.D. degree in control theory and control engineering from the Institute of Automation, Chinese Academy of Sciences (IACAS), Beijing, China, in 2022. 

He is currently an associate professor with the College of Information and Electrical Engineering, China Agricultural University, Beijing, China. His current research interests include robotic fish and intelligent control systems.
\end{IEEEbiography}

\vspace{-1cm}
\begin{IEEEbiography}[{\includegraphics[width=1in,height=1.25in,clip,keepaspectratio]{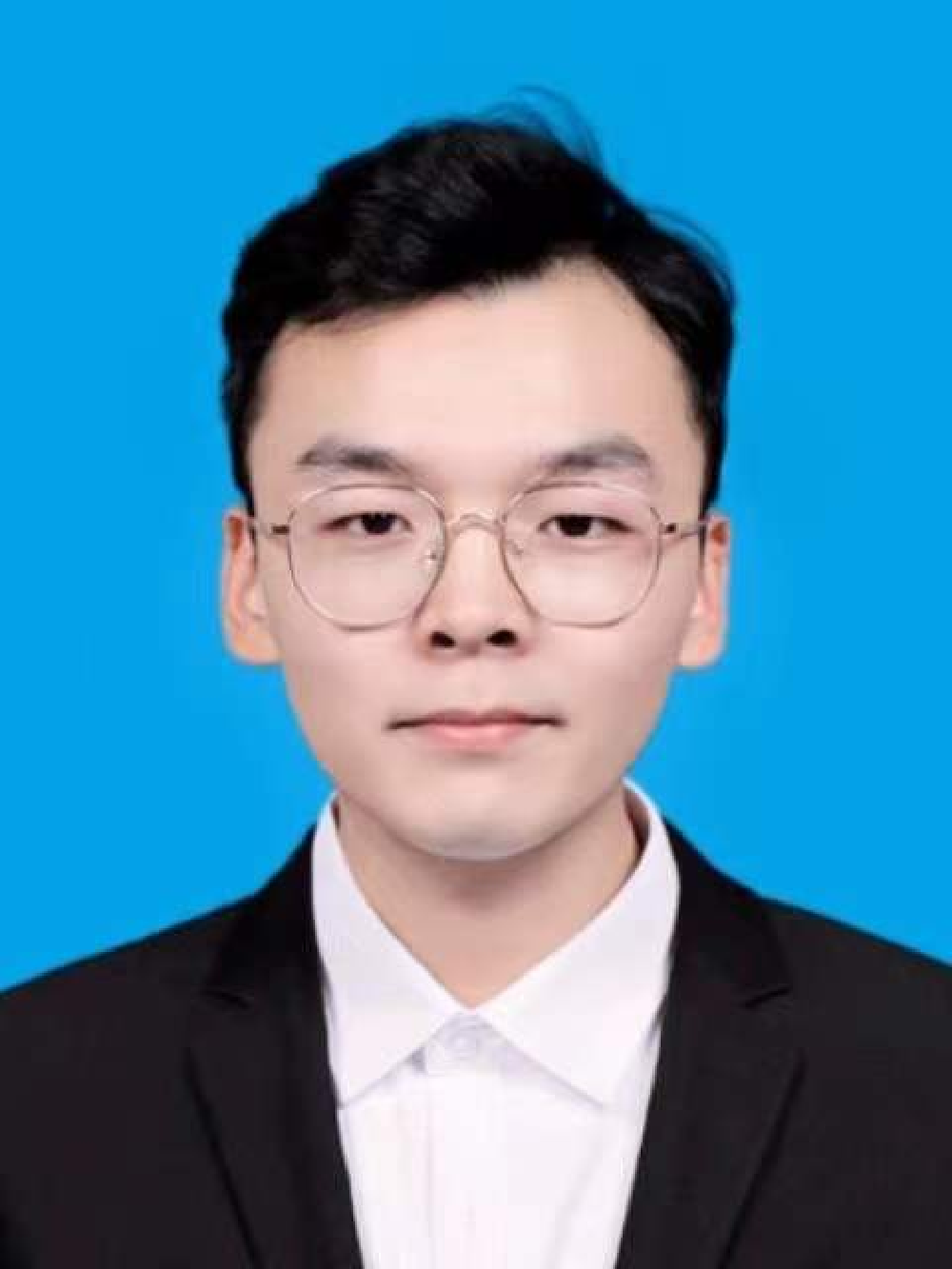}}]
{Shihan Kong}
received the B.E. degree in automation from the School of Control Science and Engineering, Shandong University, Jinan, China, in 2016 and the Ph.D. degree in control theory and control engineering from the Institute of Automation, Chinese Academy of Sciences (IACAS), Beijing, China, in 2021. 

He is currently an assistant research fellow with the College of Engineering, Peking University, Beijing, China. His current research interests include underwater robotics and underwater robotic vision.
\end{IEEEbiography}

\vspace{-1cm}
\begin{IEEEbiography}[{\includegraphics[width=1in,height=1.25in,clip,keepaspectratio]{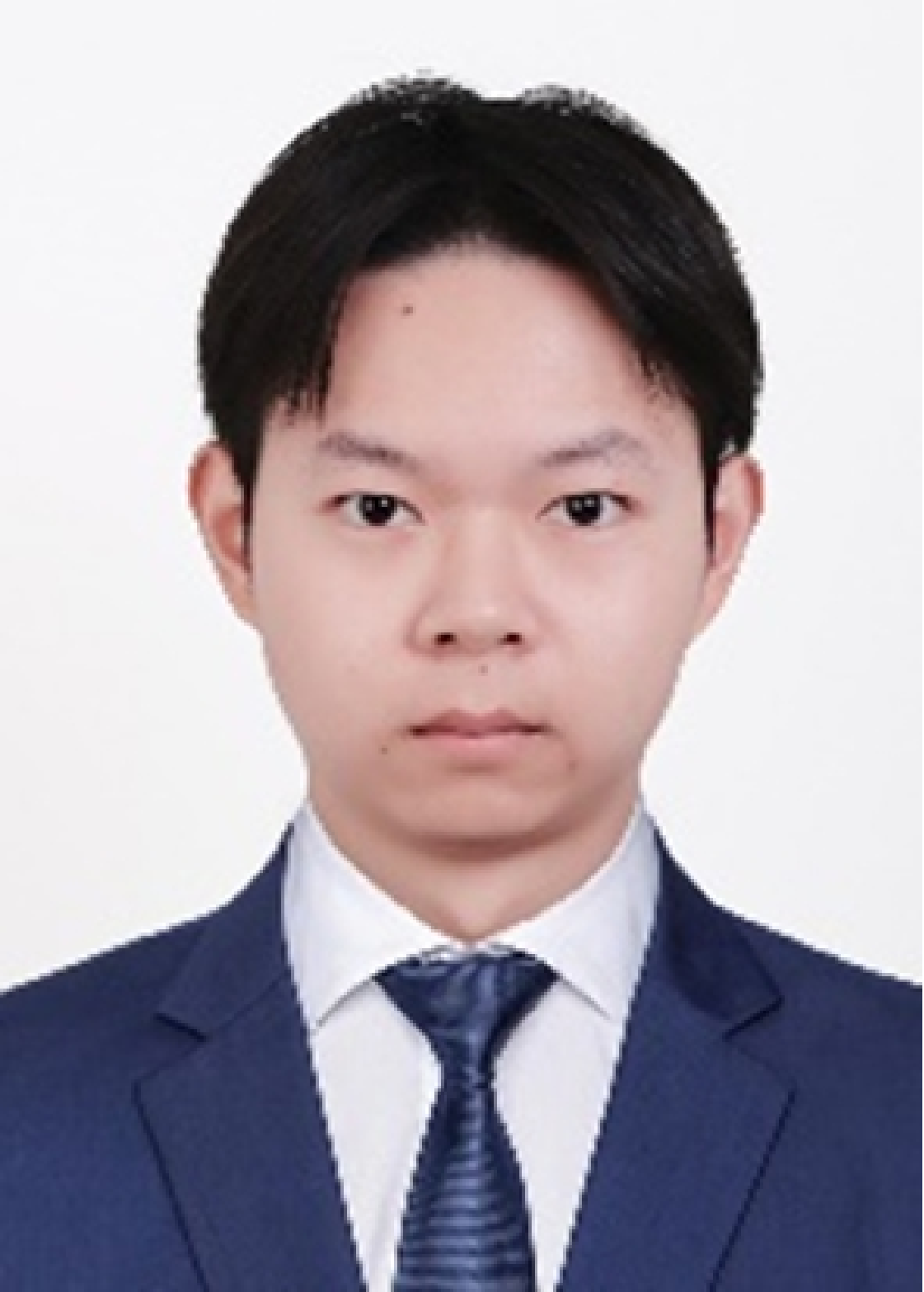}}]
{Yan Xiong}
received the B.E. degree in Mechanical Design, Manufacturing and Automation from College of Engineering, Huazhong Agricultural University, Wuhan, China, in 2022. He is currently working toward the M.E. degree in Mechanical Engineering with College of Engineering, Peking University, Beijing, China. His current research interests include bioinspired underwater robots.
\end{IEEEbiography}

\vspace{-1cm}%
\begin{IEEEbiography}[{\includegraphics[width=1in,height=1.25in,clip,keepaspectratio]{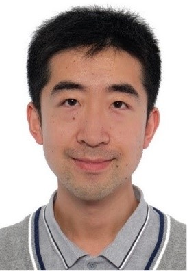}}]
{Chong Zhang}
received his B.S. in Electrical Engineering from Shandong University, China, in 2009, his M.S. and Ph.D. in Electronic and Computer Engineering from Hong Kong University of Science and Technology in 2010 and 2017 respectively. He then joined Tencent Robotics X in 2018 as a senior researcher. His research interest span computational neuroscience, developmental robotics, and AI-based computer animation.
\end{IEEEbiography}

\vspace{-1cm}
\begin{IEEEbiography}[{\includegraphics[width=1in,height=1.25in,clip,keepaspectratio]{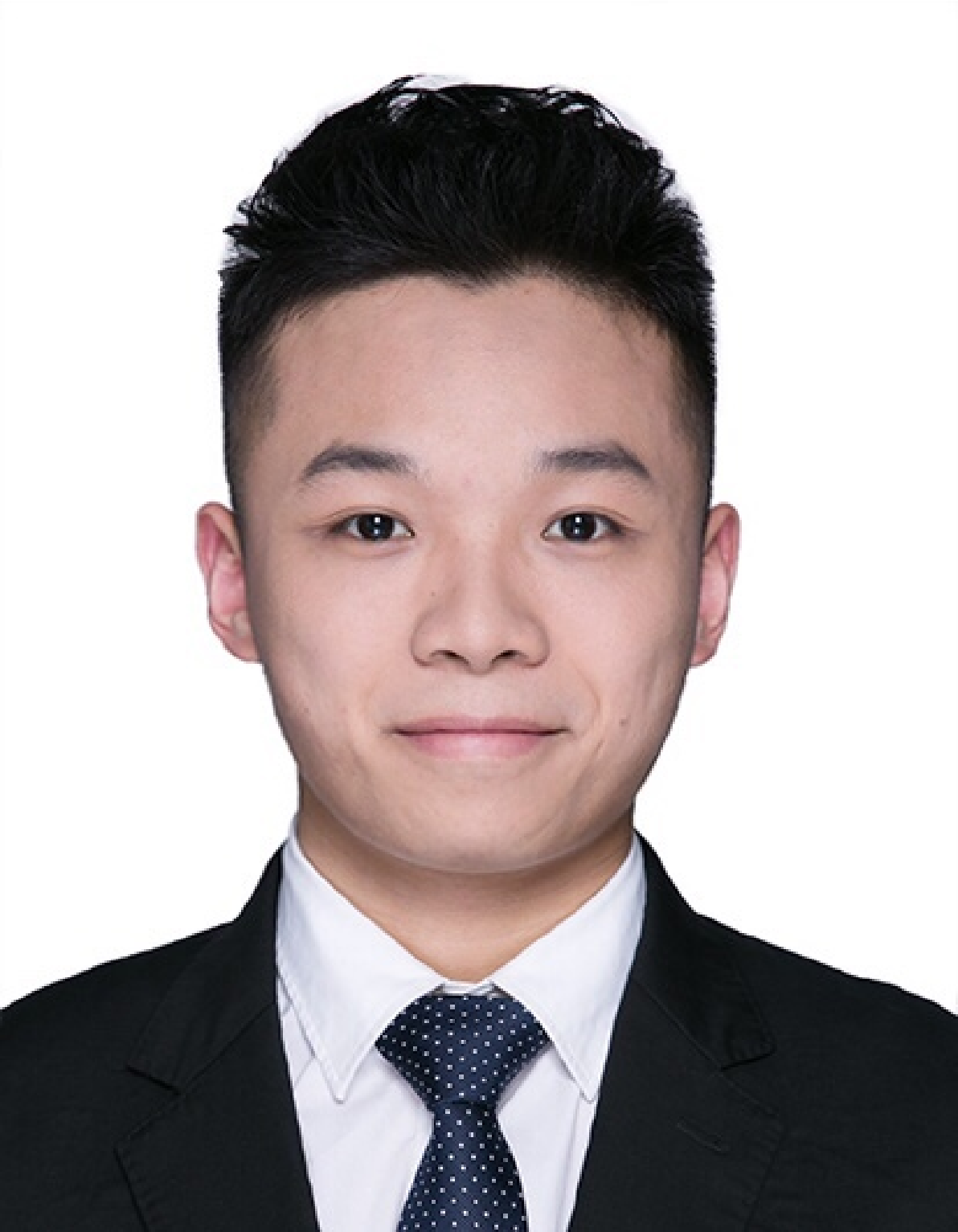}}]
{Yuzhen Liu}
received his B.S. degree from Huazhong University of Science \& Technology, Wuhan, China, in 2017, and Ph.D. degree from Tsinghua University, Beijing, China, in 2022. He is currently a Senior Researcher with the Tencent Robotics X Lab. He is interested in SLAM, robotics, and embodied AI.
\end{IEEEbiography}

\vspace{-1cm}
\begin{IEEEbiography}[{\includegraphics[width=1in,height=1.25in,clip,keepaspectratio]{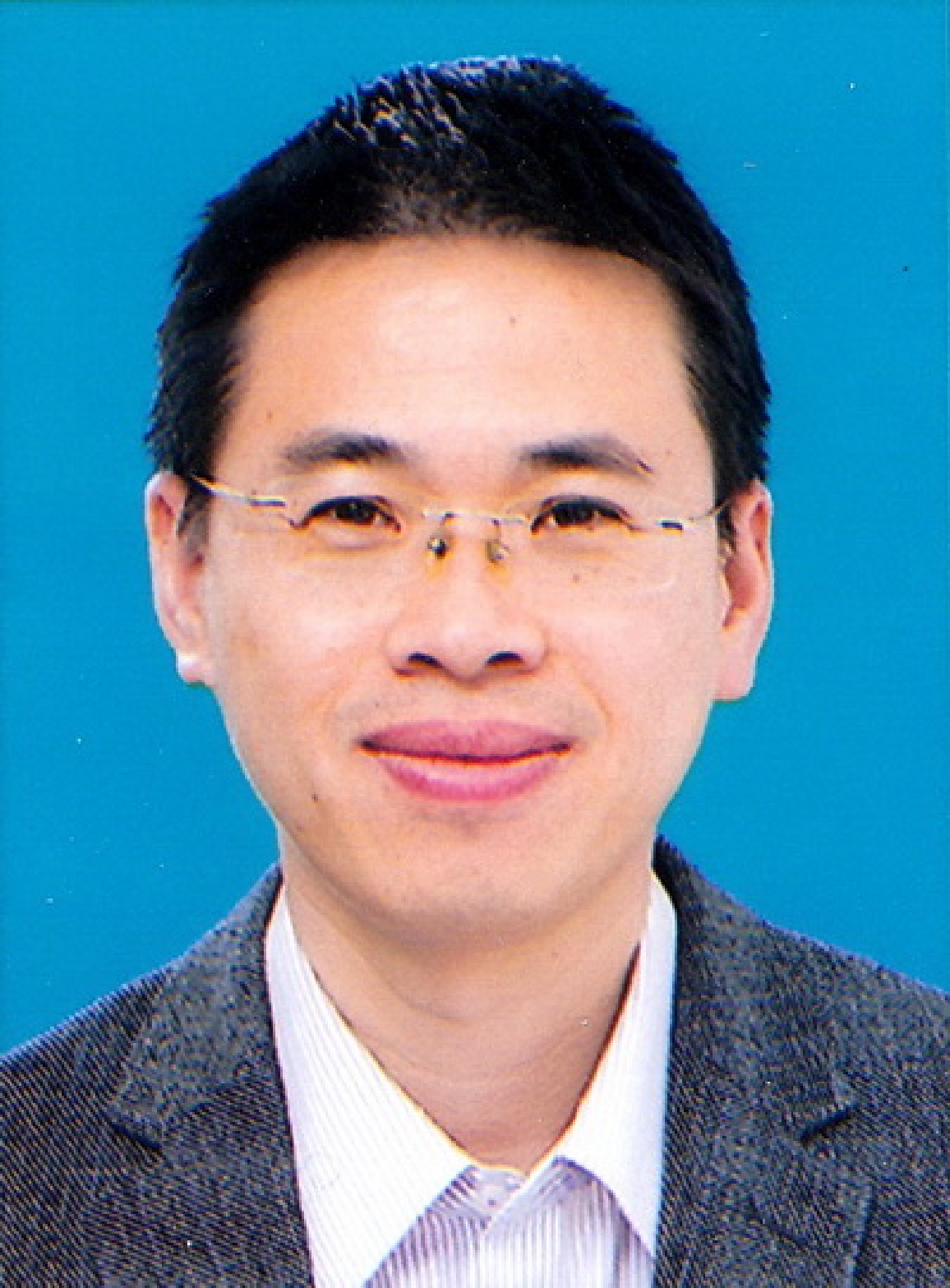}}]
{Junzhi Yu (Fellow, IEEE)}
received the B.E. degree in safety engineering and the M.E. degree in precision instruments and mechanology from the North University of China, Taiyuan, China, in 1998 and 2001, respectively, and the Ph.D. degree in control theory and control engineering from the Institute of Automation, Chinese Academy of Sciences, Beijing, China, in 2003.

From 2004 to 2006, he was a Postdoctoral Research Fellow with the Center for Systems and Control, Peking University, Beijing. In 2006, he was an Associate Professor with the Institute of Automation, Chinese Academy of Sciences, where he became a Full Professor in 2012. In 2018, he joined the College of Engineering, Peking University, as a Tenured Full Professor. His current research interests include intelligent robots, motion control, and intelligent mechatronic systems.
\end{IEEEbiography}

\end{document}